\documentclass[conference]{IEEEtran}
\IEEEoverridecommandlockouts
\usepackage{cite}
\usepackage{amsmath,amssymb,amsfonts}
\usepackage{algorithmic}
\usepackage{graphicx}
\usepackage{textcomp}
\usepackage{xcolor}
\usepackage{caption}
\usepackage{multirow}
\usepackage[hidelinks]{hyperref}
\usepackage{booktabs}
\usepackage{subcaption}
\def\BibTeX{{\rm B\kern-.05em{\sc i\kern-.025em b}\kern-.08em
    T\kern-.1667em\lower.7ex\hbox{E}\kern-.125emX}}

\begin{document}
\let\oldtwocolumn\twocolumn

\title{Spectral Signature Mapping from RGB Imagery \\for Terrain-Aware Navigation
}

\author{%
\IEEEauthorblockN{Sarvesh Prajapati}
\IEEEauthorblockA{\textit{Northeastern University}\\
Boston, USA\\ prajapati.s@northeastern.edu}
\and
\IEEEauthorblockN{\hspace*{20mm}Ananya Trivedi}
\IEEEauthorblockA{\hspace*{20mm}\textit{Northeastern University}\\
\hspace*{20mm}Boston, USA\\ \hspace*{20mm}trivedi.ana@northeastern.edu}
\and
\IEEEauthorblockN{\hspace*{20mm}Nathaniel Hanson}
\IEEEauthorblockA{\hspace*{20mm}\textit{Massachusetts Institute of Technology}\\
\hspace*{20mm}Lexington, USA\\ \hspace*{20mm}nhanson2@mit.edu}

\and
\IEEEauthorblockN{\phantom{Pad}}
\IEEEauthorblockA{\phantom{\textit{Affil}}\\ \phantom{City}\\ \phantom{mail}}
\and
\IEEEauthorblockN{\phantom{Pad}}
\IEEEauthorblockA{\phantom{\textit{Affil}}\\ \phantom{City}\\ \phantom{mail}}
\and
\IEEEauthorblockN{\hspace*{-55mm}Bruce Maxwell}
\IEEEauthorblockA{\hspace*{-55mm}\textit{Northeastern University}\\
\hspace*{-55mm}Seattle, USA\\ \hspace*{-55mm}b.maxwell@northeastern.edu}
\and
\IEEEauthorblockN{\hspace*{-15mm}Ta\c{s}k{\i}n Pad{\i}r$^1$}
\IEEEauthorblockA{\hspace*{-15mm}\textit{Northeastern University}\\
\hspace*{-15mm}Boston, USA\\ \hspace*{-15mm}t.padir@northeastern.edu}
\thanks{${^1}$Ta\c{s}k{\i}n Pad{\i}r holds concurrent appointments as a Professor of Electrical and Computer Engineering at Northeastern University and as an Amazon Scholar. This paper describes work performed at Northeastern University and is not associated with Amazon.}
}

\maketitle

\begin{abstract}

Successful navigation in outdoor environments requires accurate prediction of the physical interactions between the robot and the terrain. 
Many prior methods rely on geometric or semantic labels to classify traversable surfaces. 
However, such labels cannot distinguish visually similar surfaces that differ in material properties. Spectral sensors enable inference of material composition from surface reflectance measured across multiple wavelength bands. Although spectral sensing is gaining traction in robotics, widespread deployment remains constrained by the need for custom hardware integration, high sensor costs, and compute-intensive processing pipelines. In this paper, we present the
RGB Image to Spectral Signature Neural Network (RS-Net), a deep neural network designed to bridge the gap between the accessibility of RGB sensing and the rich material information provided by spectral data. RS-Net predicts spectral signatures from RGB patches, which we map to terrain labels and friction coefficients. The resulting terrain classifications are integrated into a sampling-based motion planner for a wheeled robot operating in outdoor environments. Likewise, the friction estimates are incorporated into a contact-force–based MPC for a quadruped robot navigating slippery surfaces. 
Overall, our framework learns the task-relevant physical properties offline during training and thereafter relies solely on RGB sensing at run time.
The code is available at \url{https://github.com/prajapatisarvesh/RS-Net}.


\end{abstract}


\section{Introduction}
Robots are increasingly deployed in everyday settings, ranging from self-driving cars~\cite{autonomousdrivingsurvey} and search-and-rescue missions~\cite{hanson2024forestbiomassmappingterrestrial} to wildfire prevention~\cite{wildfire1,wildfire2}. In such unstructured environments, reliable autonomy demands more than obstacle avoidance. It requires precise reasoning about how terrain properties influence motion. For example, vehicles must modulate braking on icy roads, and off-road platforms should bypass dense swamps to avoid entrapment. These scenarios show that perception must move beyond geometry and semantics toward reliable estimates of robot–terrain interactions.

Several off-road motion planning pipelines use RGB cameras to identify terrain from images~\cite{materialcnn,ganav}. In some cases, visually similar surfaces with different physical properties, such as ice on asphalt, may be mislabeled, leading to invalid traversability cost maps. Depth cameras and LiDAR are often used to estimate the ease of motion over a surface~\cite{evora,trip}. However, the robot must first drive the terrain to create a dataset, which risks hardware damage and necessitates tuning specific to the operating site.

In contrast, spectral sensors offer a non-invasive way to estimate material properties. This is accomplished by leveraging distinct patterns of light absorption and reflection, known as spectral signatures, to characterize underlying material composition. These capabilities are finding use in robotics applications such as wildfire risk monitoring~\cite{hanson2024forestbiomassmappingterrestrial}, manipulation~\cite{erickson2019classification, slurp}, and exploration~\cite{hansonprospect}. By mapping spectral signatures to physical quantities such as moisture content, rigidity, or surface type, the same sensing stack can be repurposed across robots and environments with minimal changes to the processing pipeline. However, challenges such as custom mounts, calibration requirements, large datasets, and high sensor costs currently limit deployment at scale.

RGB cameras are inexpensive, widely available, and already standard in robotic perception pipelines such as object detection~\cite{yolo} and tracking~\cite{trackingeverything}. Compared to hyperspectral systems, they are cheaper, lighter, and more power efficient, which simplifies integration on mobile platforms. Advances in deep learning  now allow RGB imagery to approximate measurements traditionally obtained from more information-dense sensors.
\begin{figure*}
    \centering
    \includegraphics[width=0.8\textwidth]{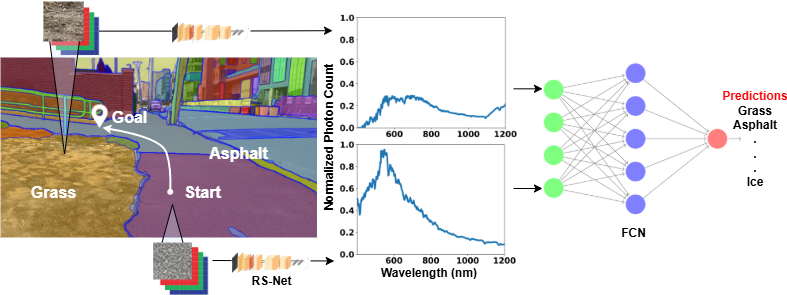}
           \label{front_fig}
           \captionof{figure}{Our approach takes RGB image patches (Left) as input and predicts their spectral signatures (Middle). These predicted spectra are then mapped to terrain classes by tuning the final fully connected layer (Right). The resulting terrain predictions inform navigation. The robot selects a path along the asphalt road, which is easier to traverse than the surrounding dense grass.}
           \label{fig:paper_intro}
\end{figure*}

Motivated by this, we seek to retain the deployment advantages of RGB cameras while recovering spectral sensor features. 
We introduce the
RGB Image to Spectral Signature Neural Network or RS-Net, a deep neural network architecture trained on spectral data collected from diverse materials. It maps RGB image patches to their corresponding spectral signatures. These estimates are passed to a lightweight feedforward neural network whose weights are fine-tuned for the target physical property. We retrain this network once per task, enabling the same neural network architecture to perform terrain classification and friction estimation. Our entire inference pipeline runs at approximately 5 Hz, making it suitable for real-time robotic applications. Fig.~\ref{fig:paper_intro} outlines the proposed architecture.

We validate our method in both simulation and hardware experiments. The terrain classification is used in a sampling-based motion planner for outdoor navigation of a skid-steer robot. Similarly, the friction estimates are integrated into a model predictive control (MPC) scheme for a quadrupedal robot operating on slippery surfaces. Finally, we also discuss how the proposed approach generalizes to other robots and additional physical properties relevant to off-road planning.

\section{Related Work}
Inaccurate estimation of the coefficient of friction (CoF) can lead to slip and skid in mobile robots, resulting in degraded motion planning performance~\cite{ours_ral,trivedi_wokshop}. Many methods that rely on RGB cameras estimate friction via terrain classification. These approaches depend heavily on the quality of segmentation masks and are vulnerable to visual aliasing and mislabeling~\cite{materialcnn}. Moreover, models trained in one region, season, or camera setup may encounter significantly different illumination, exposure, or texture statistics at deployment, leading to poor generalization. Recent retrieval-augmented vision–language models avoid the need for paired image–friction labels by using language supervision for CoF estimation~\cite{vlmgronav}. However, this comes at the cost of increased inference time. To address these limitations, we bypass semantic segmentation entirely and infer the friction coefficient from predicted spectral signatures. Our approach employs lightweight convolutional and fully connected networks, enabling inference at approximately 5 Hz. This supports real-time deployment while avoiding the computational overhead associated with retrieval-based language models.
\begin{figure*}[!t]
    \centering
    \includegraphics[width=0.85\linewidth]{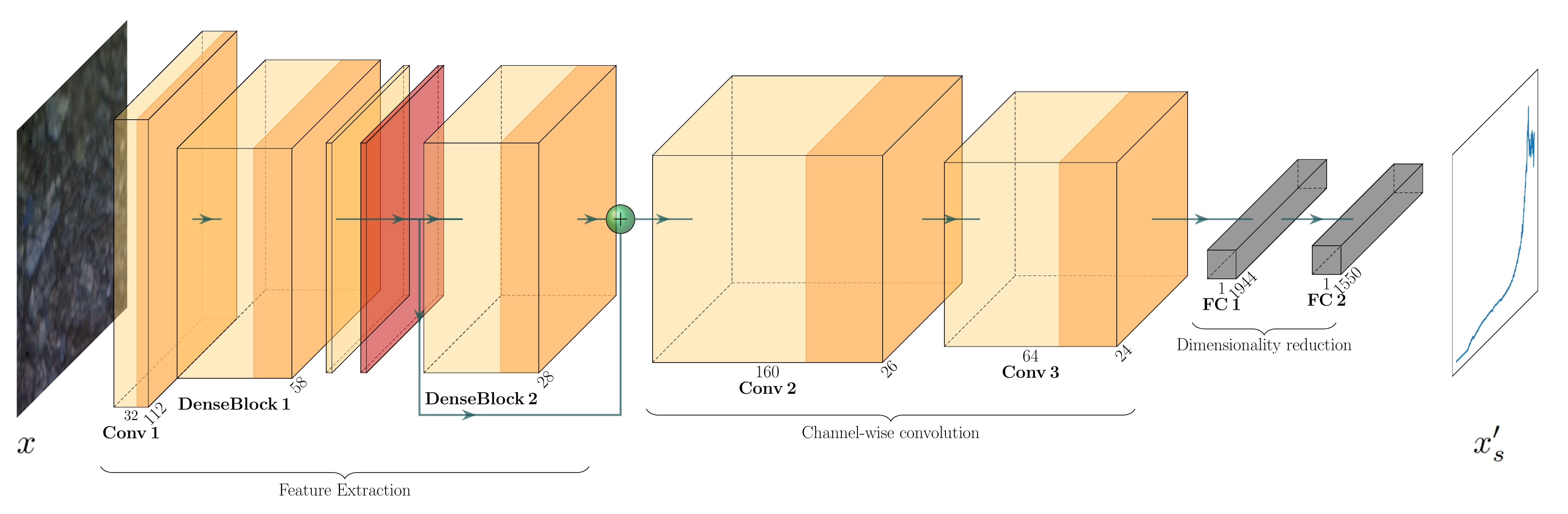}
    \caption{Architecture to predict spectral signature $x_s'$ from RGB image $x$. }
    \label{fig:rsnet}
\end{figure*}
Thermal sensing has been explored for friction estimation by identifying material categories under challenging lighting conditions~\cite{deepthermal}.  Its practicality is limited by a short operating range of less than 0.5 m and a low frame rate, which constrain deployment on mobile platforms. Heat transfer tactile sensors estimate material properties through physical contact, which can lead to sensor wear or pose a risk of damage to the robot~\cite{touchthermal}. Acoustic sensing has been applied to surface classification by recording wheel–ground interaction sounds, but it remains highly susceptible to ambient noise~\cite{microphone}. Wheel encoder and IMU signals are commonly used to estimate robot pose through Kalman filtering frameworks~\cite{ananyaicra24,amco,smppi}. These approaches rely on motion-induced signals to capture terrain interaction but require platform and terrain specific calibration, as well as sufficient on-ground excitation. This limits their generalization to new environments and robot configurations. In contrast to these approaches, our method relies solely on RGB input, offering sufficient sensing range for mobile robotic applications. It is fully non-invasive, eliminating the risk of wear or damage associated with contact-based sensors. Estimating material properties directly from spectral signatures removes the need for terrain-specific tuning and supports deployment across varied environments with no additional calibration.

Spectral and hyperspectral sensors capture material-specific reflectance by measuring how surfaces respond across a broad range of wavelengths~\cite{hsisurvey}. This enables fine-grained discrimination between materials that appear visually similar~\cite{slurp,vastdata}. However, real-world deployment faces several challenges~\cite{hanson2024forestbiomassmappingterrestrial}. These include high sensor cost, complex mechanical integration on mobile robots, and large memory requirements due to the high-dimensional nature of the underlying data. We address the limitations of spectral sensing by predicting spectral profiles directly from RGB terrain patch. This removes the need for expensive, high-bandwidth sensors while leveraging compact, low-cost RGB cameras already integrated into most robotic platforms. The reduced data dimensionality enables efficient storage and real-time inference, making the approach practical for large-scale, field-deployable systems.

Recent traversability estimation methods are beginning to move beyond pure geometry by incorporating richer terrain cues. TERP~\cite{terp} constructs stability-based costs from elevation maps, while TRIP~\cite{trip} predicts traversability from steppability analysis on dense 3D reconstructions. However, both these methods lack explicit modeling of material properties and can still misclassify low-friction surfaces. EVORA~\cite{evora} learns traction distributions from exteroceptive sensing and incorporates these estimates into a Conditional Value at Risk (CVaR)-aware motion planner. However, it relies on supervised learning from physical interaction data collected over rough terrain, which limits its scalability. In contrast, we propose a unified approach that estimates planner-agnostic material properties by fine-tuning a lightweight fully connected head on top of spectral inference network. This allows the same backbone to be reused across motion planning tasks with minimal retraining.

\section{Methodology}
In this section, we first outline the neural network architecture used for mapping RGB inputs to spectral signatures. We then describe the output head that maps the predicted spectra to motion planning relevant quantities such as terrain classification and coefficient of friction. Subsequently, we explain the learning objectives and the optimization strategy used in the training process. Finally, we describe how the resulting predictions are integrated into existing motion-planning pipelines for both wheeled and legged robots.

\subsection{RS-Net Architecture and Task-Specific Prediction Head}
As shown in Fig.~\ref{fig:rsnet}, RS-Net consists of three main stages: (i) feature extraction with DenseNet, (ii) feature fusion and channel reduction via shallow CNN layers, and (iii) spectral projection through fully connected layers. We explain each of these components below.
\subsubsection*{DenseNet for Feature Extraction}
To extract global features from the input RGB image \(x\), we use a pretrained deep neural network, DenseNet-169~\cite{densenet}. Other DenseNet variants can be appropriately substituted, as the first two dense blocks share the same structure across different versions of DenseNet. We provide a high-level overview of the data flow through DenseNet below.

We use $G(\cdot)$ to denote generic convolution–pooling operations, which appear both in the initial stem and in transition layers within the network.  Each dense block is denoted by \( H(\cdot) \), where the \(\ell\)th layer receives the feature maps of all preceding layers, \(\mathbf{x}_0, \ldots, \mathbf{x}_{\ell-1}\), as input:
\begin{equation}
    \mathbf{x}_\ell = H_\ell([\mathbf{x}_0, \mathbf{x}_1, \ldots, \mathbf{x}_{\ell-1}])
\end{equation}
Here, \([\mathbf{x}_0, \mathbf{x}_1, \ldots, \mathbf{x}_{\ell-1}]\) denotes the concatenation of the feature maps produced by layers \(0\) through \(\ell - 1\).

The image $x$ first passes through the operation $G(\cdot)$, which produces the initial feature map.
\begin{equation} \label{eq:conv_and_pooling}
    x_1 = G(x)
\end{equation}
\(x_1\) is now passed sequentially through the first two dense blocks of the DenseNet backbone. The first block, which contains 6 convolutional layers, captures fine-grained local textures and low-level visual features. The second block, consisting of 12 convolutional layers, builds upon these features to extract higher-level contextual information.

In our implementation, the first dense block outputs $x_7$, which is passed through a transition layer denoted $G(\cdot)$ to produce $x_8$. This intermediate representation is then processed by the second dense block, whose output is $x_{21}$. The data flow through these operations is summarized as follows:
\begin{align}
    x_7 &= H_7([x_1, x_2, ..., x_6]) \\
    x_8 &= G(x_7) \\
    x_{21} &= H_{21}([x_8, x_9, ..., x_{20}])
\end{align}

\begin{figure*}[t]
    \centering
    \includegraphics[width=0.75\linewidth]{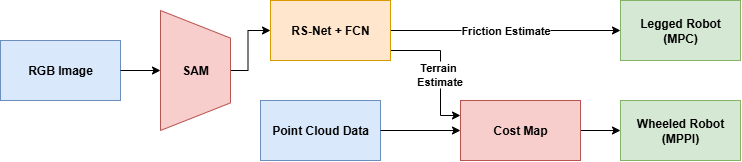  }
    \caption{Overview of our motion planning pipeline. Segment Anything (SAM) extracts image patches from the RGB input, which are passed through RS-Net followed by a Fully Connected Network to predict per-patch friction estimates or terrain labels. These predictions inform friction constraints or terrain-aware cost maps for downstream motion planning algorithms.}
    \label{fig:rs_net_framework}
\end{figure*}
We construct a fused global feature representation by concatenating two intermediate outputs:the max-pooled output from the first transition layer and the features produced by the second dense block. Let $\oplus$ denote channel-wise concatenation. The resulting fused tensor is given by:
\begin{equation}
    x_f = x_8 \oplus x_{21}
\end{equation}

\subsubsection*{CNN Layers}
The fused feature map \( x_f \) is processed by two lightweight convolutional layers for dimensionality reduction. The first layer reduces the channel count from 160 to 64 and downsamples the spatial resolution. The second layer further compresses the representation from 64 to 9 channels, producing a compact descriptor that both regularizes the model and reduces the computational cost of the subsequent spectral projection.

\subsubsection*{Fully Connected Layers}
The 9-channel feature map is flattened and passed through two fully connected layers. The first projects the flattened input from 1944 to 1550 dimensions, followed by a second fully connected layer that maintains the dimensionality at 1550. Denoting this fully connected transformation by the composite mapping \( S(\cdot) \), the final spectral output is given by:
\begin{equation}
    x'_s = S(x_f)
\end{equation}

\subsubsection*{Task-Specific Prediction Head}
We append a lightweight fully connected network (FCN) after the spectral projection to map the predicted spectral profile \(x'_s\) to the desired planner-facing quantity, e.g., material class or friction coefficient. The spectral vector \(x'_s \in \mathbb{R}^{B_s}\) is high-dimensional, where we use \(B_s = 1550\) bands. To reduce this to a compact task-specific representation, we employ a multi-layer perceptron (MLP) with the structure \(1550 \rightarrow 512 \rightarrow 128 \rightarrow K\), where \(K\) is the output dimension. For classification tasks, \(K\) denotes the number of classes and a softmax activation is applied to produce per-class probabilities. For regression tasks, \(K=1\) to output a scalar value such as surface friction. All intermediate layers use Gaussian Error Linear Unit (GELU) activation and dropout for regularization.


\subsection{Neural Network Training Workflow}

Training is conducted end-to-end, allowing gradients from the final task to flow back and influence the intermediate spectral representation learned by RS-Net. This coupling ensures that the DenseNet features are shaped not just to minimize reconstruction error but also to benefit the downstream task.

At inference time, only RGB input is provided. RS-Net predicts the spectral vector $x'_s$, which is then processed by the task-specific Fully Connected Network (FCN) to yield the required physical property. These predictions are made patch-wise across the image. The resulting patch-level outputs are then separately post-processed into a full-resolution output map, aligned with the input image.

We train on the VAST dataset~\cite{vastdata}, which contains high-resolution RGB images of terrain surfaces, paired spectral reflectance profiles, and 9-DoF IMU data across 11 terrain types under varied lighting conditions. RS-Net is trained on six terrain classes: asphalt, brick, grass, ice, sand, and tile. It is also evaluated on five held-out classes: carpet, concrete, gravel, mulch, and turf, to assess generalization.

We first pretrain RS-Net to learn an RGB-to-spectrum mapping using a mean-squared error loss. Given the ground-truth spectrum $x_s \in \mathbb{R}^{B_s}$ and the RS-Net prediction $x'_s\in \mathbb{R}^{B_s}$, the training loss is:
\begin{equation}
    \mathcal{L}_{\text{spec}} = \frac{1}{B_s} \left\| x'_s - x_s \right\|_2^2
\end{equation}
After pretraining, the task-specific FCN maps the predicted spectrum $x'_s$ to a planner-facing property. The full architecture is then trained jointly using a weighted combination of spectral and task losses:
\begin{equation}
    \mathcal{L} = \alpha \mathcal{L}_{\text{task}} + (1 - \alpha) \mathcal{L}_{\text{spec}}, \quad \text{with } \alpha = 0.7
\end{equation}

For classification tasks such as terrain labeling, $\mathcal{L}_{\text{task}}^{\text{terrain}}$ is defined as the standard cross-entropy loss between the ground-truth label $y_i$ and the predicted softmax probability ${y'}_i$ over $K$ terrain classes. The ground-truth labels $y_i$ are obtained from~\cite{vastdata}:
\begin{equation}
    \mathcal{L_{\text{task}}^{\text{terrain}}} = -\sum_{i=1}^Ky_i\log({y'_i})
\end{equation}

For scalar regression tasks such as estimating a friction coefficient, $\mathcal{L}_{\text{task}}^{\text{friction}}$ is defined as the mean absolute error (L1 loss) between the predicted and ground-truth values ${z'}_n$ and $z_n$, over a batch of $N$ samples. The ground-truth friction values $z_n$ are obtained from~\cite{brandaofriction}:
\begin{equation}
\mathcal{L}_{\text{task}}^{\text{friction}} = \frac{1}{N} \sum_{n=1}^{N} \left| {z'}_n - z_n \right|
\end{equation}

We train for 50 epochs using the Adam optimizer~\cite{adamopt} with a learning rate of $1 \times 10^{-3}$ on a Lenovo Legion 5 Pro laptop with an Intel i7 12700H processor, 32 GB RAM, and an NVIDIA RTX 3070 GPU. Despite using a consumer-grade, lightweight GPU, the full training process completed in under 10 hours.

\subsection{Integrating RS-Net into the Navigation Pipeline}
To integrate RS-Net into the navigation pipeline, we preprocess RGB frames into material-specific patch inputs. Segmentation masks are first obtained using Segment Anything (SAM)~\cite{kirillov2023segment} to isolate homogeneous regions. From each mask, we extract the largest square crop and apply minor recropping to standardize dimensions, ensuring compatibility with RS-Net’s fixed input size prior to inference. An overview of the perception-to-planning pipeline is shown in Fig.~\ref{fig:rs_net_framework}, illustrating how RS-Net predictions are converted into actionable inputs for downstream navigation in both wheeled and legged robots.

\section{Experiments and Results}\label{sec:experiments}
We perform three sets of experiments to evaluate RS-Net. First, we assess its ability to reconstruct spectral signatures from RGB images. Next, we integrate the friction prediction module into a quadruped robot’s control stack and evaluate its impact on navigation performance. Finally, we deploy RS-Net’s terrain classification outputs on a wheeled robot to guide outdoor navigation.

\begin{figure}
    \centering
    \includegraphics[width=0.95\linewidth]{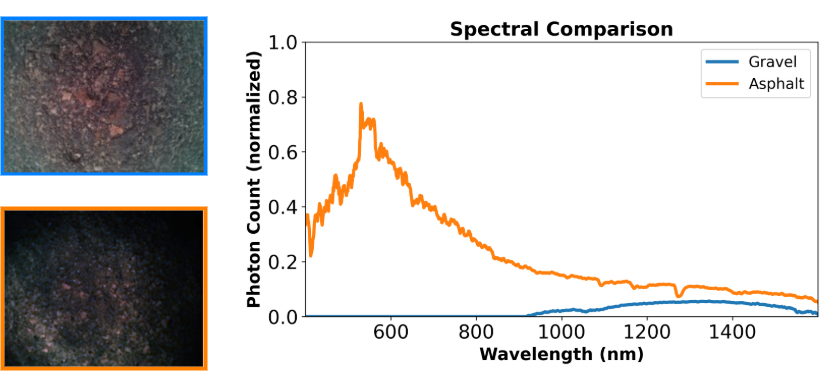}
    \caption{Top: RGB patch of gravel. Bottom: RGB patch of asphalt. RS-Net predicts distinct spectral signatures for these visually similar surfaces.}
    \label{fig:texture-similartiy}
\end{figure}

\subsection{Evaluation of RGB-to-Spectral Mapping}

\begin{table}[h]
  \centering
  \caption{MSE for Spectral Profile Prediction.}
  \label{tab:comp}
  \begin{tabular}{|c|c|}
    \hline
    Model & Score \\
    \hline
    SRCNN & 0.10$\pm$0.01 \\
    \hline
    U-Net Regression & 0.03$\pm$0.01 \\
    \hline
    \textbf{RS-NET} & \textbf{0.002$\pm$0.001}\\
    \hline
  \end{tabular}
  
  
\end{table}
A key motivation for our approach is to address the failure of appearance based terrain classification methods in distinguishing between materials that look similar but differ in physical properties. Fig.~\ref{fig:texture-similartiy} illustrates this using two terrain patches, asphalt and gravel, that appear nearly identical in RGB imagery. Despite this visual similarity, RS-Net reconstructs distinct spectral signatures for each. Asphalt exhibits consistently higher photon counts than gravel, allowing clear separation in the spectral domain. This demonstrates that RS-Net learns to extract fine-grained visual cues beyond texture to infer material composition. This capability is essential for downstream motion planning tasks that require reliable reasoning about terrain properties in visually ambiguous environments.

To compare spectral signature prediction performance, we used 30\% of the VAST dataset as the validation set. We evaluated our method against two popular networks: SRCNN~\cite{SRCNN} and U-Net~\cite{unet}, where we removed their final layers and added fully connected heads to perform spectral regression. As seen in Table~\ref{tab:comp} RS-Net achieves lower MSE across all terrain classes compared to both baselines.



\begin{figure}[thb]
    \centering
    \includegraphics[width=0.3\linewidth]{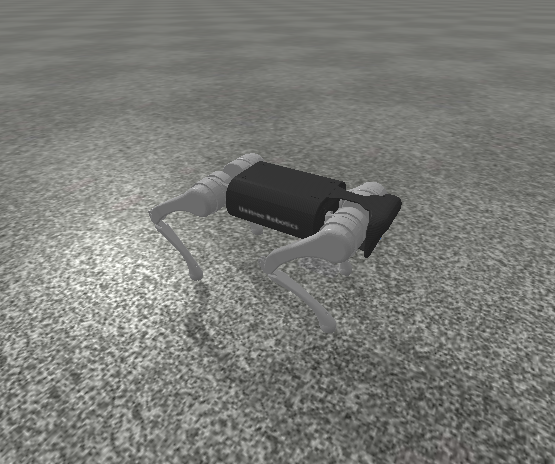}
    \includegraphics[width=0.3\linewidth]{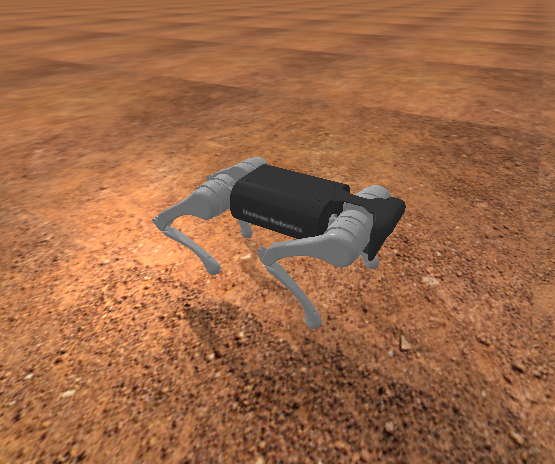}
    \includegraphics[width=0.3\linewidth]{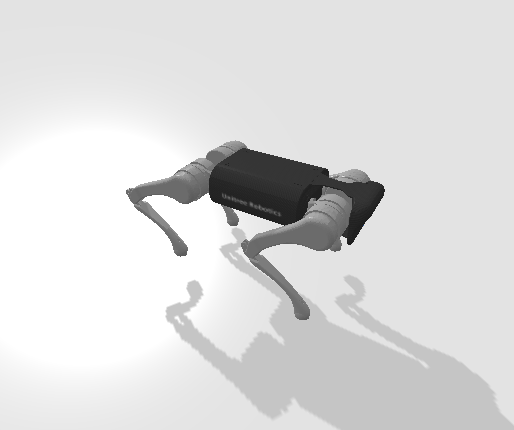}
    \caption{Representative surface textures used in simulation. From left to right: asphalt, brick, and ice.}
    \label{fig:sim-monte}
\end{figure}

\subsection{Quadrupedal Robot Navigation}
As shown in Fig.~\ref{fig:rs_net_framework}, our method directly predicts the coefficient of friction \(\mu\) from RGB input. We put this to test by deploying the resulting estimates in a quadruped robot navigating slippery surfaces. To prevent foot slippage, the ground reaction forces must satisfy the friction cone constraint:

\begin{equation}
    \sqrt{f_x^2 + f_y^2} \leq \mu f_z
    \label{eq:friction-cone}
\end{equation}
where \(f_x\), \(f_y\), and \(f_z\) are the contact forces along the respective axes. The shape of this cone, and therefore the feasible set of forces, is directly defined by \(\mu\)~\cite{mit_cheetah}, making friction estimation critical for stable locomotion. The Linear Model Predictive Controller (LMPC) we use is based on~\cite{ours_ral}. The MPC parameters used for simulation and hardware experiments are shown in Table~\ref{tab:mpc_parameters}. The controller tracks a desired center of mass height of 0.32 m and a forward velocity of 0.5 $m/s$, while enforcing the friction cone constraint during optimization. We evaluate our method in both simulation and hardware. In each case, RGB images are segmented into patches using SAM. A friction estimate is computed for each patch, and the minimum predicted value is used to inform the LMPC.

\begin{table}[h]
    \centering
    \caption{LMPC parameters used for quadruped navigation.}
    \label{tab:mpc_parameters}
    \footnotesize
    \begin{tabular}{lc}
    \toprule
    \textbf{Parameter} & \textbf{Value} \\
    \midrule
    Stepping Frequency & 2.75\,Hz \\
    Foot Height & 0.10\,m \\
    Planning Horizon & 10 steps \\
    Planning Timestep & 0.05\,s \\
    Position Weight (Z) & 50 \\
    Velocity Weights (X, Y) & 10, 5 \\
    Angular Velocity Weights (X, Y, Z) & 0.2, 0.2, 1.0 \\
    Roll and Pitch Weights & 0.2, 0.2 \\
    Control Penalty Weights & 1e-6 \\
    \bottomrule
    \end{tabular}
\end{table}

\subsubsection{Monte Carlo Evaluation in Simulation}
We use PyBullet ~\cite{coumans2021} to simulate 1000 episodes for rigorous evaluation of our framework across varied terrains. In each trial, one terrain class is sampled from a predefined library consisting of asphalt, brick, grass, ice, sand, and tile. The corresponding surface texture is applied as the ground plane in the simulator, as shown in Fig.~\ref{fig:sim-monte}.
 A corresponding true friction coefficient \(\mu \in [0.05, 1.0]\) is set based on the material. The rendered RGB image of the terrain is passed to the trained network to estimate \(\hat{\mu}\), which is then used by the controller to perform locomotion. We compare controller performance under two conditions: one using a fixed friction value of \(\mu = 0.5\), and the other using the estimated friction \(\hat{\mu}\) predicted from RGB input. The goal is to evaluate whether learned friction estimates yield better locomotion performance across diverse surfaces. Success Rate is defined as the percentage of trials where the center of mass height remains above 0.25 meters throughout the episode. Average Slippage Ratio is computed as the ratio of tangential to vertical contact force magnitudes, with higher values indicating greater slippage. Normalized Tracking Cost measures how accurately the controller follows the desired center of mass trajectory. Normalized Effort Cost reflects the total magnitude of contact forces applied, with higher values indicating less efficient control.
\begin{table}[h]
    \caption{Results from 1000 Monte-Carlo Simulations.}
    \label{tab:montecarlo}
    \centering
    \footnotesize
    \setlength\tabcolsep{3pt} 
    \begin{tabular}{cccccc}
        \toprule
        \textbf{Metric} & \textbf{Gait} & \textbf{LMPC} & \textbf{LMPC w/ RS-Net} \\
        \midrule
        \multirow{1}{*}{\textbf{Success Rate (\%)} $\uparrow$} & Trot & 79 & 100\\
        {\textbf{Average Slippage Ratio} $\downarrow$} & Trot & 0.20 & 0.12\\{\textbf{Normalized Tracking Cost} $\downarrow$} & Trot & $1.4\pm0.3$ & $1.1\pm0.05$\\{\textbf{Normalized Effort Cost} $\downarrow$} & Trot & $1.6\pm0.2$ & $1.1\pm0.05$\\
        \bottomrule
    \end{tabular}
\end{table}

\begin{figure}[h]
    \centering
    \includegraphics[width=0.85\linewidth]{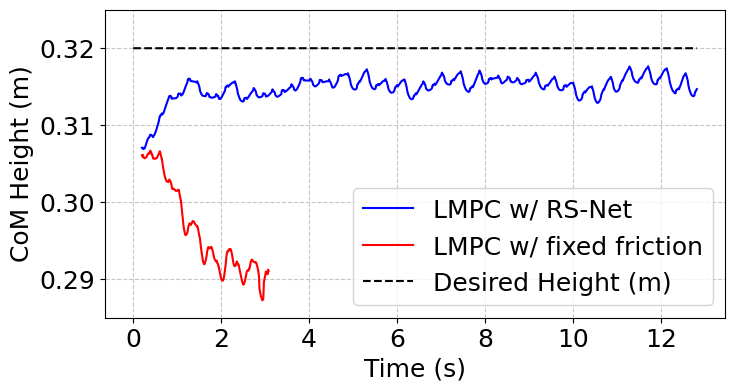}
    \caption{Height tracking experiment on slippery ice surface.}
    \label{fig:com_plot}
\end{figure}

As shown in Table~\ref{tab:montecarlo}, for the trotting gait, our framework performs better than simply using a fixed but small friction value. This improvement comes from the controller's ability to adapt the friction cone based on the predicted terrain properties, rather than applying the same force limits across all surfaces. To further evaluate stability under low-friction conditions, we test the controller on an icy surface while tracking the center of mass height at 0.32 meters, as shown in Fig.~\ref{fig:com_plot}. With active friction estimation via RS-Net, the robot maintains stability throughout the trial, whereas the fixed-friction baseline loses balance and falls within 3 seconds.

\subsubsection{Real-World Evaluation on Slippery Terrain}
We conduct hardware experiments using a Unitree Go1 quadruped traversing an indoor track that includes a deliberately low friction section, created by applying cooking oil to a whiteboard surface. All trials are performed from a fixed initial pose under consistent lighting conditions. 

\begin{figure}[t]
    \centering
    \begin{subfigure}[t]{0.72\linewidth}
        \centering
        \includegraphics[width=\linewidth]{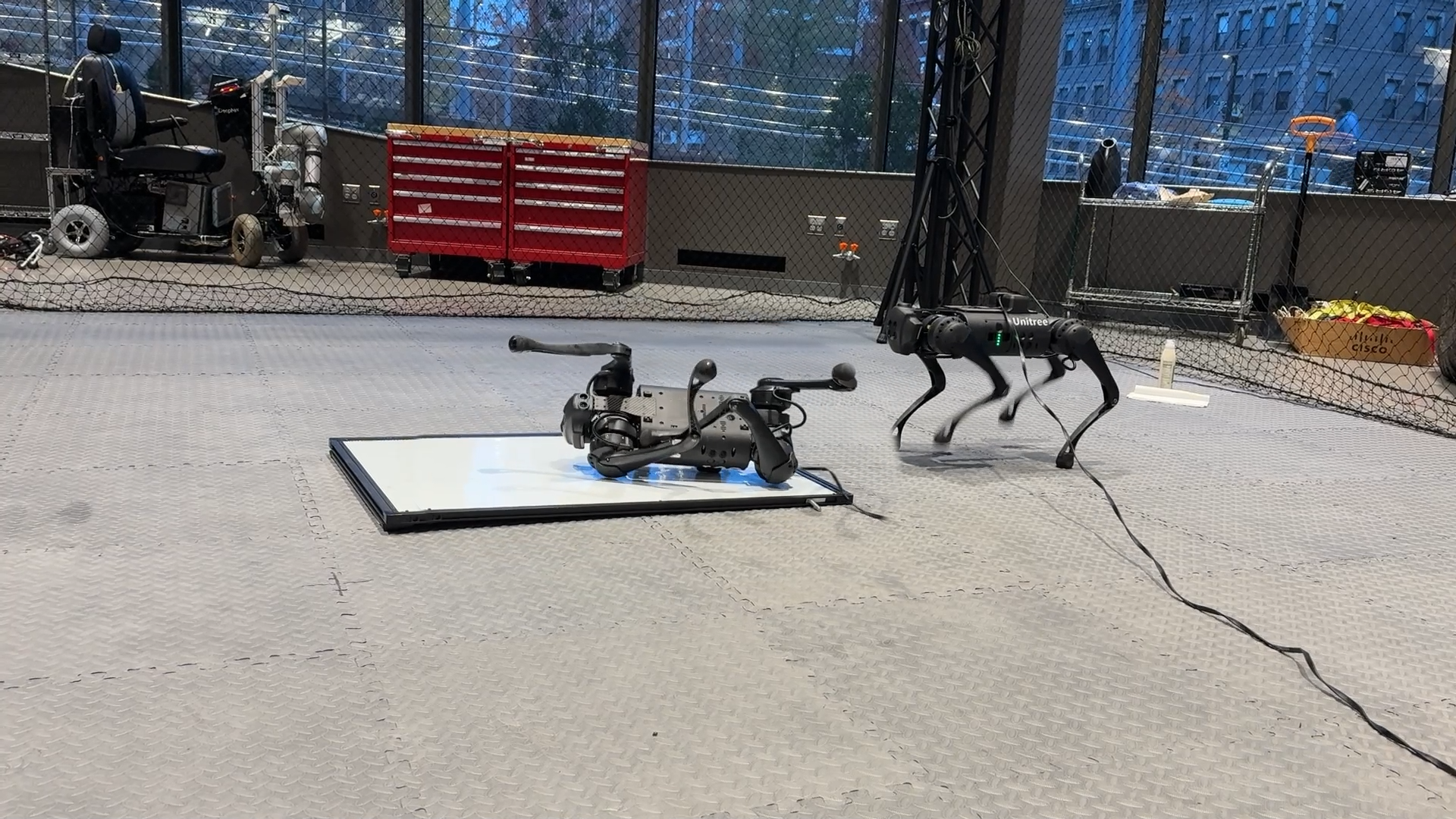}
        \label{fig:quad_fail}
    \end{subfigure}
    \hfill
    \begin{subfigure}[t]{0.72\linewidth}
        \centering
        \includegraphics[width=\linewidth]{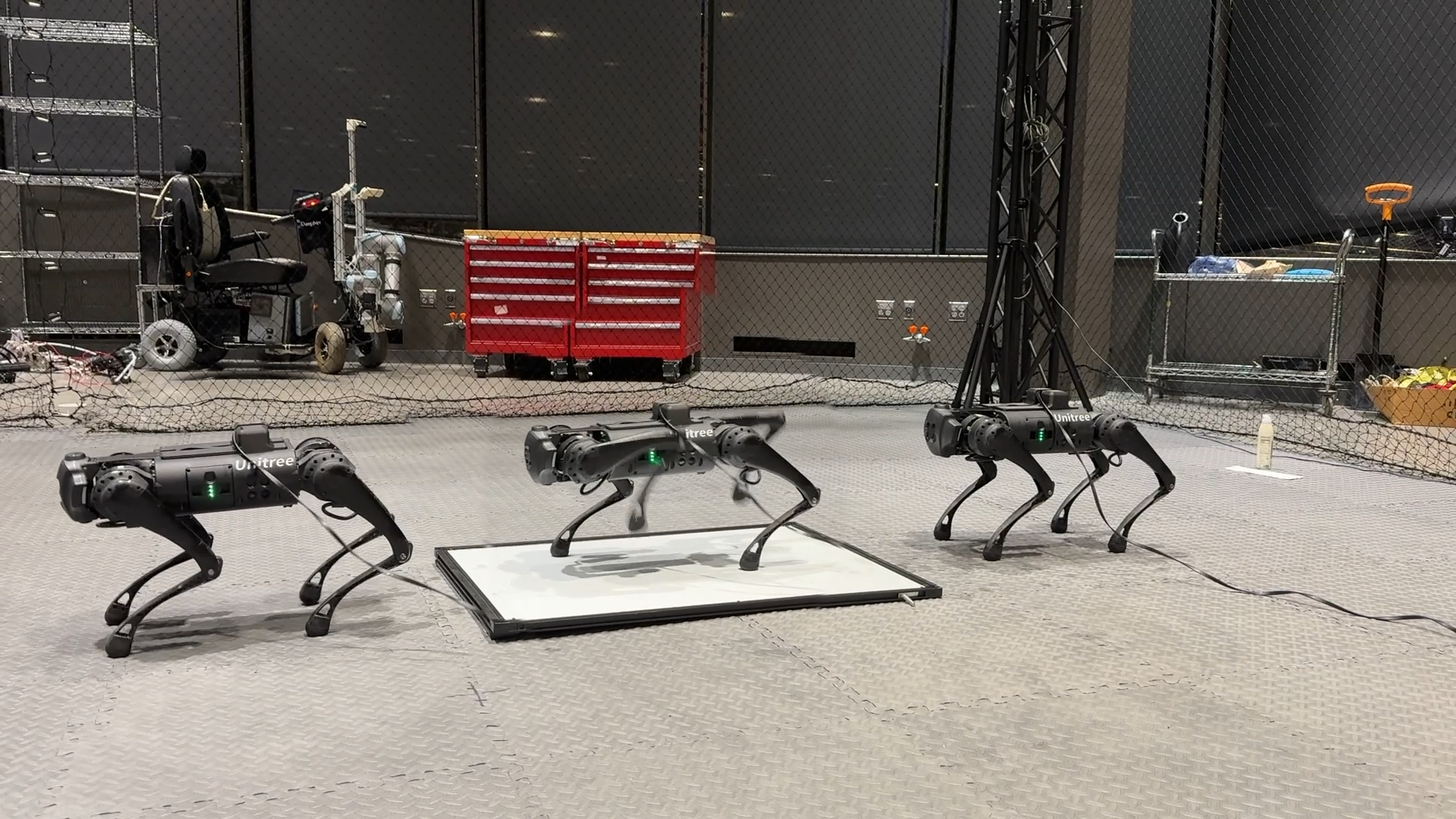}
        \label{fig:quad_success}
    \end{subfigure}
    
    \caption{Quadruped navigating from a high-friction rubber patch onto a low-friction oiled surface and back to rubber. Top: Controller with small but constant friction coefficient fails. Bottom: RS-Net-enabled online friction estimation stabilizes locomotion across changing surfaces.}
    \label{fig:quad_comparison}
\end{figure}

As shown in Fig.~\ref{fig:quad_comparison} we evaluate two controller variants: one using a constant friction value, and the other using RS-Net for real time friction estimation. In the fixed friction case, the robot slips and fails on the low friction patch. RS-Net instead predicts higher friction on regular surfaces and lower friction on the slippery region, allowing the controller to adjust its force constraints accordingly and maintain stability.


\subsection{Wheeled Robot Navigation}

\begin{figure}
    \centering
    \includegraphics[width=0.8\linewidth]{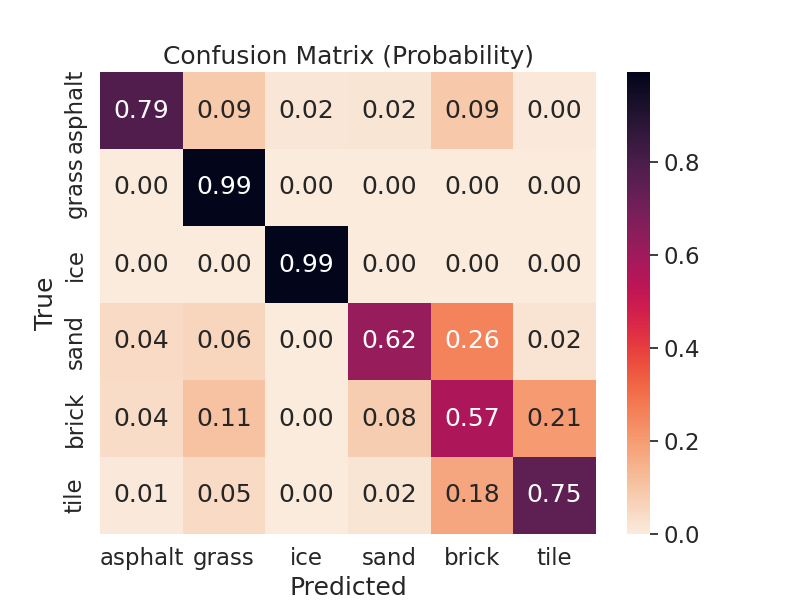}
    \caption{Confusion matrix for terrain classification}
    \label{fig:spectral_asphalt_conf}
\end{figure}

In off road navigation, certain surfaces such as asphalt are more favorable for traversal than others, such as dense grass. To account for this, we use the same RS-Net architecture as in the quadruped experiments, but fine tune only the final fully connected layer to perform terrain classification. The network outputs a discrete terrain label for each input patch, which is then used to construct a terrain aware cost map.

To assess how well RS-Net performs terrain classification, we evaluate it on the VAST validation set and visualize the results in Fig.~\ref{fig:spectral_asphalt_conf}. While several terrain classes such as ice, grass, and tile are predicted with high accuracy, others like sand and brick show moderate confusion due to similarly shaped but spatially offset spectral profiles~\cite{oursworkshop}.

Once terrain labels are predicted, they are projected into 3D space using camera–LiDAR calibration~\cite{camlidarcalib}, and ground level patches are extracted using Patchwork~\cite{lim2021patchwork}. RS-Net runs at 5 Hz, updating terrain estimates in real time. The resulting terrain aware cost map is integrated into a sampling based motion planner based on Model Predictive Path Integral (MPPI) control~\cite{smppi}. MPPI evaluates 1024 candidate trajectories per planning cycle and selects the optimal one using a weighted sum of all trajectories.

\begin{figure}
    \centering
        \includegraphics[width=0.7\linewidth]{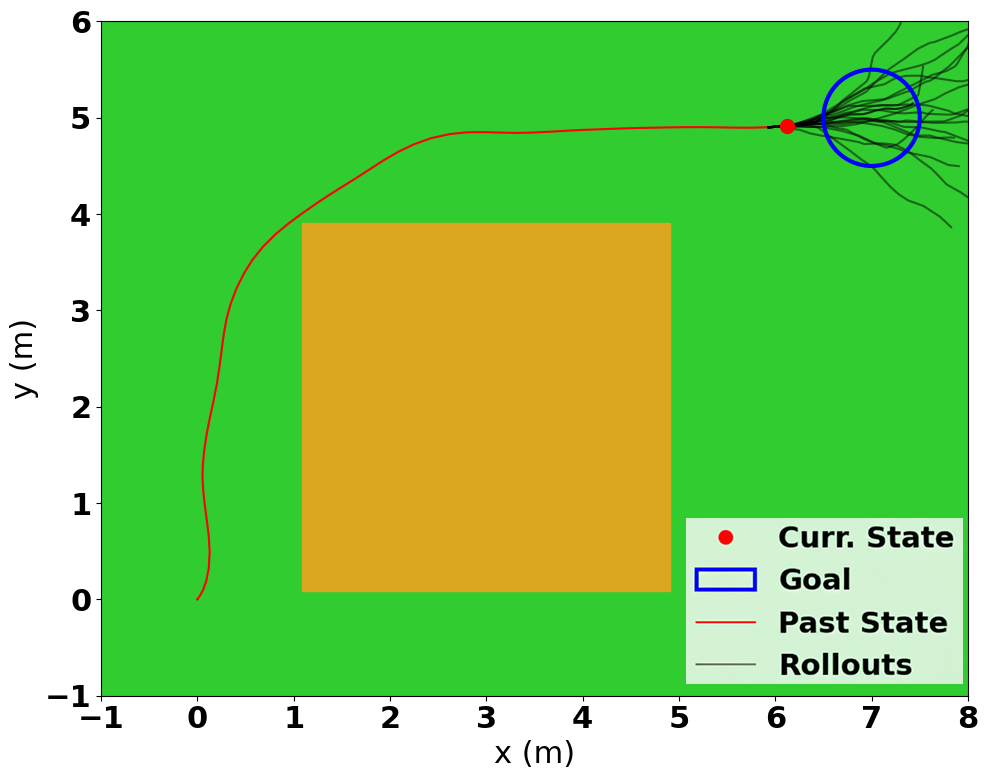}
    \caption{Grid world simulation used to tune MPPI costs. Green indicates traversable space, brown denotes high-cost material patch, and black lines show sampled rollouts.}
    \label{fig:grid-world}
\end{figure}

As shown in Fig.~\ref{fig:grid-world}, we use a 2D gridworld environment to tune the MPPI cost weights. A square material patch is randomly placed between a fixed start and goal location and treated as a high-cost region in the motion planner. The planner operates with a 5 second planning horizon, a timestep of \(\Delta t = 0.05\) seconds, and control limits of 2 $m/s$ linear velocity and 3 $rad/s$ angular velocity.

We deploy our framework on a Clearpath Jackal robot in an outdoor setting with the start and goal placed across an asphalt route interspersed with grass patches. An onboard RGB camera streams terrain images to RS-Net, which assigns higher traversal costs to grass compared to asphalt.

\begin{figure}[t]
    \centering
    \includegraphics[width=0.76\linewidth]{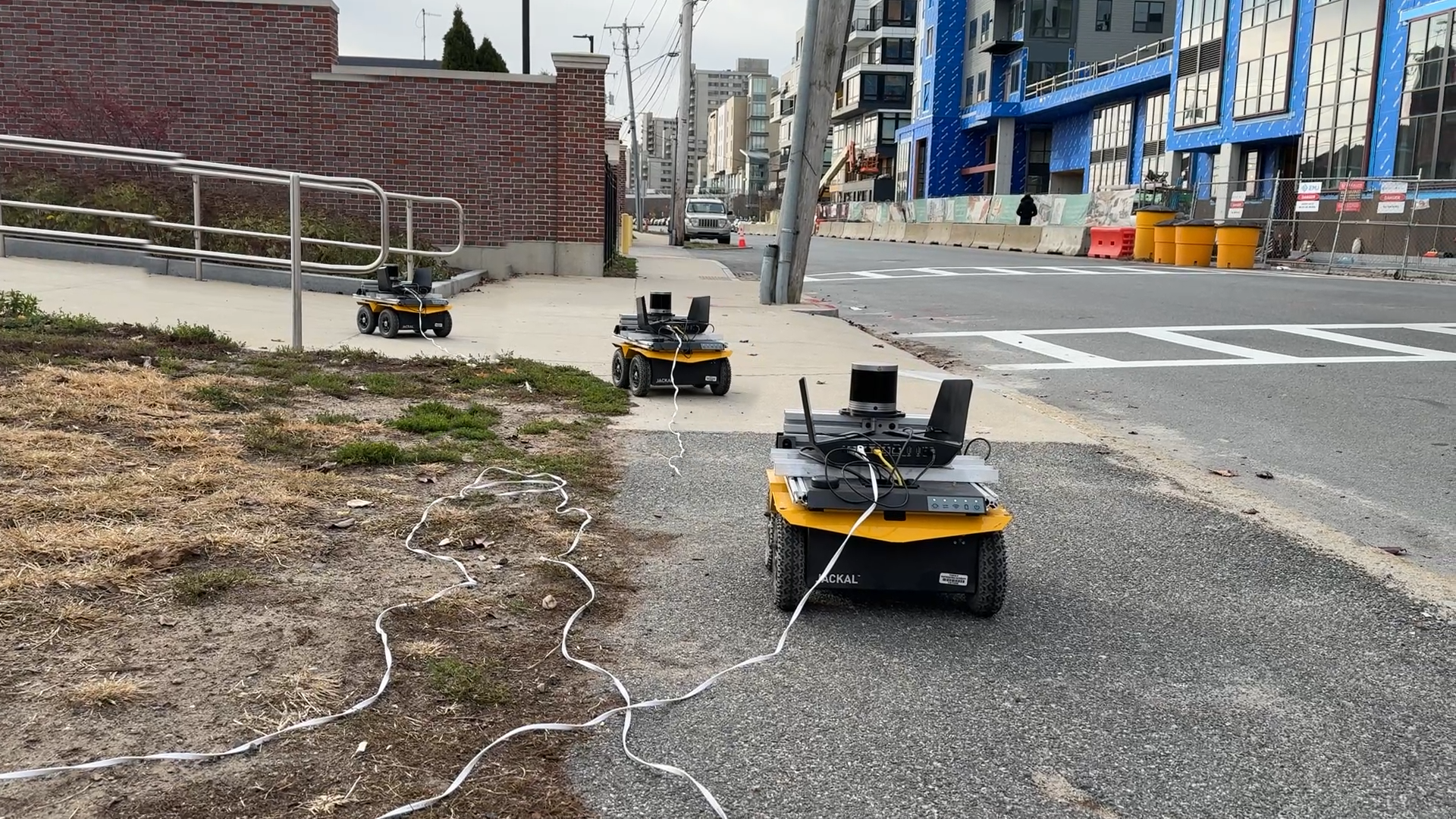}
    \caption{The RS-Net-enabled wheeled robot navigates along the asphalt path, actively avoiding the grass region to reduce traversal risk.}
    \label{fig:jackal}
\end{figure}

We compare two planning conditions: (i) Baseline MPPI, which uses geometry-only costs and ignores material cues, and (ii) RS-Net + MPPI, which integrates a terrain-aware cost layer based on RS-Net prediction. Both planners use identical MPPI hyperparameters. The baseline selects the shortest diagonal path through grass, minimizing time-to-goal but traversing less reliable terrain. In contrast, as shown in Fig.~\ref{fig:jackal}, the RS-Net-enabled planner avoids grass entirely, favoring longer asphalt-only routes. This yields comparable time-to-goal while significantly reducing the risk of getting stuck.

\section{Limitations and Future Work}

\begin{figure}
    \centering
        \centering
        \includegraphics[width=0.7\linewidth]{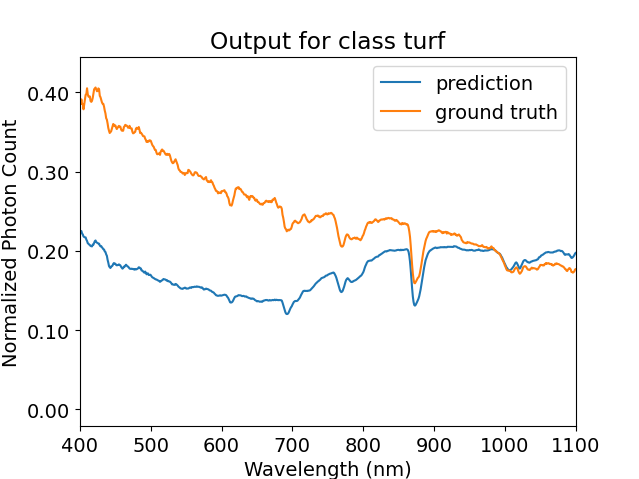}
    \caption{Predicted spectral profile v/s ground truth data for a class not in the training set.}
    \label{fig:untrained-class}
    
\end{figure}

Fig.~\ref{fig:untrained-class} reports the spectral signature predicted by RS-Net for turf terrain, which was not included in the training set. The overall spectral shape is preserved, but the predicted magnitude is consistently biased.
To mitigate this, we plan to investigate vision encoders with richer feature representations, such as Vision Transformer~\cite{beit}, and train on larger and more diverse datasets~\cite{newspec}. 

Instead of relying on a single patch, prediction accuracy may benefit from fusing information across multiple patches or viewpoints. This approach can help the model better capture the bidirectional reflectance distribution function (BRDF), which encodes how material properties vary with view and illumination. By incorporating such fine-grained physical cues, the model may reduce the discrepancy between predicted and ground truth spectra.

\section{Conclusion}

We present a modular architecture for estimating terrain properties to support motion planning in mobile robots. This is achieved by training a neural network to predict spectral signatures directly from RGB images, enabling low cost sensing without the need for dedicated spectral hardware. A simple training mechanism allows the output to be adapted for different physical properties, such as terrain class and friction coefficient. In future work, we plan to extend this framework beyond navigation to tasks such as manipulation, and to infer additional physical properties relevant to robot–environment interaction, including moisture content and terrain compliance~\cite{landclass}.

\bibliographystyle{IEEEtran}
\bibliography{IEEEabrv,references}

\end{document}